\documentclass[conference]{IEEEtran}
\IEEEoverridecommandlockouts    % to override locked commands

\usepackage{multirow}
\usepackage{lipsum}
\usepackage{amsmath}
\usepackage{amssymb}
\usepackage[ruled,vlined]{algorithm2e}
\usepackage{graphicx}
\graphicspath{{./Figures/}}
\title{\LARGE \bf Image Quality Dependent Degradation for AI Systems}

\author{
	\parbox{\textwidth}{%
		\centering
		Yannick Kees$^{1}$, Elena Hoemann$^{1}$, Frank Köster$^{1}$ , Sven Hallerbach$^{1}$
	}%
	\thanks{$^{1}$German Aerospace Center (DLR) Institute for AI Safety and Security
		{\tt\small {surname.lastname}@dlr.de, }}%
}

\usepackage{stfloats}

\usepackage{tikz}
\usepackage{float}
\usepackage{xcolor}
\usepackage[normalem]{ulem} % für \uline etc.
\usepackage{censor}

\usepackage{hyperref}
\usepackage{cleveref}
\def\etal{\emph{et al}. }
\usepackage[pagewise]{lineno}

\begin{document}
\maketitle
\begin{abstract}
Perception is one of the primary applications where neural networks outperform conventional algorithms. One example is AI systems for automated driving, which can detect pedestrians based on image data and avoid them accordingly. %\note{Auch hier wuerde ich das selbstbewusster formulieren. Das ist nicht one problem, sondern eines, wenn nicht das Hauptproblem für AD. Die Perzeption muss mit KI gemacht werden und zuverlässig sein. }
A substantial challenge with these AI systems is that their output depends heavily on the quality of the input images. For example, if an image is of inferior quality due to heavy contamination, such as noise or darkness, accurate predictions are hardly feasible. Additionally, various types of errors can occur, each with varying relevance to the trustworthiness of the underlying AI system. In particular, it may be more critical not to detect an existing person than to detect a person where there is none. Therefore, we want to show that we can still avoid the most critical errors in situations of inferior image quality. To achieve this, we aim to establish a fail-degraded system by lowering the network's confidence threshold based on the estimated image quality, enabling it to detect objects more cautiously in uncertain situations. Additionally, we present a novel method for estimating the quality of incoming images by comparing them to the training data using normalizing flows. We will also conduct experiments applying our method to state-of-the-art object detection.
In summary, we will present a design strategy for AI-based systems in automated driving that can deal with poor-quality input data without resorting to fallback solutions. Such measures enhance trust in AI-based systems and lead to an increased provision of the AI component.\ \\
GitHub: \url{https://github.com/DLR-KI/information-quality-monitor}
\end{abstract}    
\section{Introduction}
\label{sec:intro}
Perception systems are an integral part of many automated systems \cite{yurtsever2020survey}. Systems based on cameras are among the most common. However, the inherently complex and non-linear relationship between raw pixel values and semantic objects often requires the use of deep neural networks to
\begin{figure}[htbp]
\input{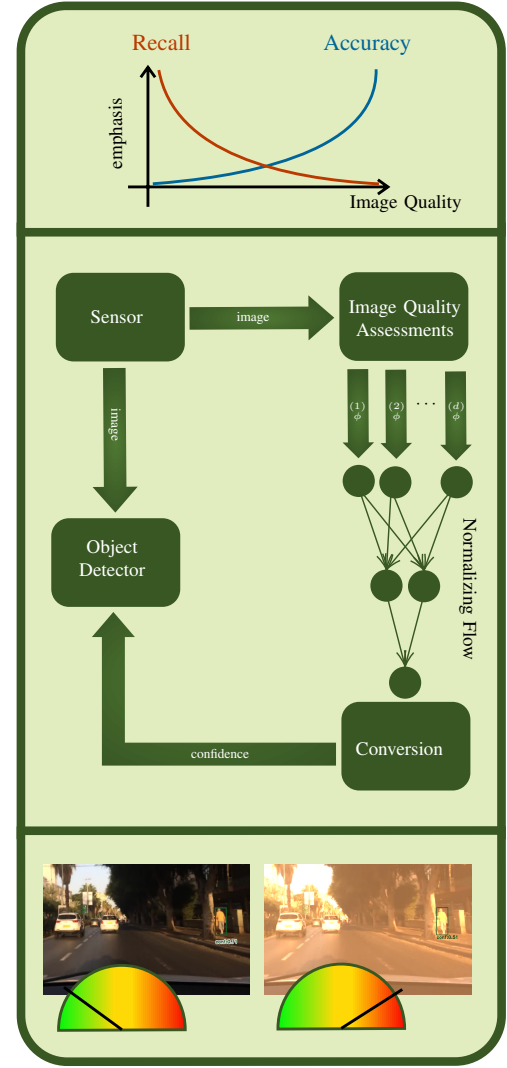}
   \caption{Overview of the fail-degraded system. For high-quality images, the network should prioritize giving the most accurate predictions, while for low-quality images, it should focus on increasing recall. See \cref{sec:method} for details.}
   \label{fig:intro}
\end{figure}\noindent
     detect objects accurately. Nevertheless, the problem with these networks is that their decision-making is challenging to understand due to the vast number of parameters they contain. This makes it difficult to argue for the safety of such systems, which is essential for their use in safety-critical applications \cite{kia2022}. 
   
In particular, it is challenging to assess the system's behavior under error-prone data. In this case, an error means corrupted input data. Such corruption can occur, for example, due to sensor wear, adverse environmental conditions, or the actions of other external actors. To ensure that such inputs cannot become a hazard, it is essential to define appropriate fault tolerance regimes for systems as well as individual subsystems \cite{Stolte2022}. At the subsystem level, fault tolerance regimes can still ensure operation at an acceptable level of risk even if a fault is present. We investigate the establishment of a fault-tolerance regime for the specific case of an AI perception system. A widely used approach for fault-tolerant systems is the development of fail-operational systems \cite{schmid2021}. These systems can continue to guarantee the full functionality of the system despite the input of corrupted data. In practice, fallback systems are often introduced that are not affected by a specific error. For our AI perception system, one example is switching from a camera-based system to a LiDAR-based system when the camera's visibility is poor. 
Instead, in this work, we aim to develop a fail-degraded system for object detection systems.

The primary purpose of a fail-degraded system is to operate with reduced functionality in the event of a fault, rather than shutting down completely or reverting to a safe state \cite{weiss2023reliabilityanalysisgracefullydegrading}. The advantage of a fail-degraded system is that it provides an extended provision of the perception system, thus increasing the trustworthiness.

If a system based on a deep learning model reaches its limits, it may encounter circumstances with which it is unfamiliar. In these situations, the accuracy may decrease. This decrease occurs because, although neural networks are very good at interpolating between data points, i.e., they perform well when the input data lies within the defined system limits, they do not extrapolate well. Hence, these networks provide poorer results at the limits, becoming less reliable in this area. So, if we are in a situation where the AI system is generally prone to error, it is crucial to be able to control this error. Since not every type of error is equally important for the trustworthiness of the subsystem, controlling means that forcing the exclusion of critical mistakes is more critical than general accuracy. On the other hand, when we are in familiar territory, we can confidently focus on delivering the most accurate results possible. This adaptive shift of focus is what we require for our fail-degraded system.

We want to examine this approach using the example of pedestrian detection in automated driving for automated emergency breaking. To achieve this, we need to identify the most critical error in this case so that we can avoid it in unfavorable situations. There are various established approaches to defining what a critical error is in this case. For example, a pedestrian who is already obscured by another pedestrian is less relevant for detection, see Feifel \etal \cite{Feifel2022}. Alternatively, for pedestrians who are far away, false positives are not as critical as those just in front of the camera, see Lee \etal \cite{Lee2024}. Taking this idea further, one can define a criticality measure by estimating the potential time for collision and also include that information during training, like Lyssenko \etal \cite{Lyssenko2022}.

In our case, we want to make the simplified assumption that overlooking a pedestrian is generally more critical than failing to detect one where there is none. While up to level 2 vehicles, false negatives detections are neglecteble for safety-consideration because the driver of the ego-vehicle must be capable of breaking on his or her own \cite{tuevwhitepaper}, we consider higher levels of automation, where false negatives lead to front collision with unprotected humans, which is more severe than a back collision due to unnecessary breaking, where the chasis of the car protect all vehicle inmates. Additionally, false negatives are silent: a positive detection can be verified by an additional sensor, but a negative detection remains unnoticed.

% EXPLAIN
For our fail-degraded system, this means that we should be more cautious when processing poor-quality input data, while we should be more confident with our outputs for high-quality input data. In general, we want to make two main contributions with this paper.
\begin{itemize}
    \item We develop a fail-degraded system for the perceptual subsystem leveraging input monitors.
    \item We are developing a new quality-aware monitoring system to trigger the fail-degraded behavior.
\end{itemize}
\section{Related work}
\label{sec:formatting}
\textbf{Fault resistance.} A commonly used approach is to make them robust against faulty data. There are two main approaches: either process the data to remove the fault, or include the faulty data in the training process to make the network more robust.

In the first case, one can consider a performance-limiting influence variable and attempt to remove its effect. For example, Fursa \etal \cite{Fursa2021} has shown that weather conditions negatively affect network accuracy. Zhang \etal \cite{Zhang2023} combine different preprocessing steps for making the weather in the image appear clearer again. Another important technique is the use of redundancies provided by additional sensors as shown by Li \etal \cite{Li2022}.

The second approach makes the network less error-prone by incorporating them into the training. Yang \etal \cite{Yang2022} show an overview of the different possibilities for augmenting the training data. he problem with these approaches is that they only work up to a certain margin of error, and that overall model performance is reduced even if the input data itself is error-free \cite{RamirezAgudelo2025}. \\
Instead of maintaining full performance, we allow reduced but controlled performance through a fail-degraded system. Many current methods for those systems monitor operational conditions using internal or external sensors \cite{Hossam2025}. In contrast, we want to set up the fail-degraded system for the perception system only to address poor image quality. For that, there is literature that addresses uncertainty in input data (aleatoric uncertainty) by adapting the loss function and network architecture of the prediction network \cite{valdenegrotoro2022deeperlookaleatoricepistemic}. In contrast, we present a method that is independent of the object detector and can be easily combined with different detector architectures as a plug-and-play solution without retraining.

\textbf{Image Quality Assessments.}
Image Quality Assessments (IQA) are qualitative evaluation criteria for image data that often aim to reflect human perception of good quality. No-reference IQA is a subclass that does not require a reference image; it only outputs a scalar value that describes image intrinsics. Thus, Mittal \etal \cite{Mittal2012} use deviations from natural scene statistics to quantify possible distortions towards a human assessment of good quality, which is measured by averaging subjective rankings, also referred to as mean opinion score (MOS). In Mittal \etal \cite{Mittal2013}, the authors exclude human judgment by comparing pure image statistics to a reference data set. As a continuation, Venkatanath \etal \cite{Venkatanath2015} provide an IQA, where no reference dataset or human opinion is needed for computation, but still correlates well to human subjective scores.
Instead of these analytical methods, deep neural networks have been increasingly used for image evaluation in recent years. For example, Kim \etal \cite{Kim2019} or Huang \etal \cite{Huang2024} show how the MOS score can be learned using convolutional networks. Liu \etal \cite{Liu2017} show how image quality can be evaluated by synthetically distorting source images and then estimating their differences.\\

\textbf{Normalizing Flows.} A Normalizing Flow (NF) is a neural network $\eta_\theta:\ \mathbb{R}^d\to\mathbb{R}^d$ that is by construction invertible. Their goal is to transform a given, possibly complex distribution into a simpler one  \cite{rezende2015variational}. One example of these networks is the real-valued volume preserving (Real-NVP) network by Dinh \etal \cite{Dinh2016}. These networks learn a transformation to a Gaussian distribution by maximizing the logarithmic likelihood given through\begin{align} \label{al:ll}
    \ell(\eta_\theta, x) =  \log\left(|\det\nabla_x \eta_\theta(x)|\right)-\frac{1}{2}\left\|\eta_\theta(x) \right\|^2- \frac{d}{2}\log(2\pi).
\end{align} Marchal \etal \cite{Marchal2019} show how to use NFs for density estimation. We want to use NFs for OOD detection of image data, since they do not require OOD data for training but learn the distribution of the existing training data directly. However, experiments such as Nalisnick \etal \cite{Nalisnick2018} show that these models have great difficulty with this task. Kirichenko \etal \cite{Kirichenko2020} argue that this is because NFs mainly learn local features of images, but not semantic information about the whole image. Another way to use NFs effectively is to train them on extracted features, e.g., from deep learning backbones, as by Cook \etal \cite{Cook2024}. In contrast this, we want to train NFs on handcrafted features.
\section{Method}
\label{sec:method}
Our fail-degraded AI system is designed to make the network respond more cautiously when it receives poor-quality input images. In cases of poor-quality images, the system should prioritize maintaining a high recall level. In contrast, for high-quality images, it should make predictions as accurately as possible. To do this, we assume that we have given an object detection algorithm that we train on an image dataset $X\subset \mathbb{R}^{n\times m \times 3}$. During deployment, this algorithm now receives additional information about image quality, in addition to its input data, and then sets the object recognition confidence threshold based on this information. The overview of the entire pipeline is shown in Figure \ref{fig:intro}. In the following subsections, we will first define how we detect the presence of a fault and then demonstrate how to utilize this information to adjust the network's confidence threshold. 
\subsection{Monitoring}
Highly automated systems must always know the exact reliability of their input data; in our case, we need to precisely evaluate and analyze the incoming image quality. By image quality, we refer to the various statistical and visual properties of an image. This means for an image $x\in\mathbb{R}^{n\times m \times 3}$ we consider different properties $\phi^{(i)}:\mathbb{R}^{n\times m \times 3}\to\mathbb{R}$ for $i=1,\dots,d$. These functions are non-reference image quality assessment criteria. These should be different in pairs and each reflect different image quality characteristics. Current state-of-the-art detection models rely heavily on features extracted from large, pretrained models such as ResNet \cite{he2016deep} or DINO \cite{caron2021emerging}. These features, however, are designed to represent only the semantic content of the image and ignore smaller augmentations, which have a significant influence on the assessment of image quality. That is why we switch to handcrafted features, which we know address different aspects of image quality and can quantify a functional insufficiency of AI Models. Examples of different assessment criteria can be found in \cref{tab:iqa}.

\begin{table}
  \caption{Overview of the selected no-reference image quality properties $\phi^{(i)}$.}
  \label{tab:iqa}
  \centering
  \begin{tabular}{@{}p{0.3\columnwidth}p{0.65\columnwidth}@{}}
  \hline
    IQA & Explanation \\
\hline
    Brightness & \hangindent=1em Mean image value in gray-scale values \\
    Colorfulness & \hangindent=1em Quantification on how shiny the colors in an image appear, see \cite{Hasler2003} \\
    Contrast & \hangindent=1em Standard deviation of pixel values in gray-scale values\\
    Energy & \hangindent=1em Integral of the image in the Fourier domain\\
    Overexposure & \hangindent=1em Ratio of high-value pixels\\
    PSNR & \hangindent=1em Logarithmic mean image value\\
    Saturation & \hangindent=1em Mean pixel saturation measured in the HSV color space \\
    Sharpness & \hangindent=1em Laplacian applied to the image\\
    SNR & \hangindent=1em Signal-to-noise ratio\\
    Skewness & \hangindent=1em Average cubic deviation from mean pixel value\\
    Spectral center & \hangindent=1em Center of mass in the fourier domain\\
    Underexposure & \hangindent=1em Ratio of low-value pixels\\
\hline
  \end{tabular}
\end{table}
We then define the overall quality of an image as 
\begin{align}
    \Phi(x)=\begin{bmatrix}\phi^{(1)}(x) ,\cdots, \ \phi^{(d)}(x) \end{bmatrix}.
\end{align}  
Now that we have defined image quality more precisely, we still need a definition of what exactly constitutes good quality. Therefore, we define the image quality as good if its properties are represented in the training data. The reason is that object recognition performs particularly well in familiar situations but struggles in unfamiliar environments, because neural networks excel at interpolation but struggle with extrapolation. In order to determine whether the quality of an image occurred in the same or a similar way in training data, we train a normalizing flow $\eta_\theta:\mathbb{R}^d\to\mathbb{R}^d$ over the image quality in the training data as a solution to the minimization problem 
\begin{align}\label{al:min}
    \min_\theta - \mathbb{E}_{x\sim X}\left[ \ell(\eta_\theta, \Phi(x))\right], 
\end{align}
where $\ell$ is the log-likelihood defined in \cref{al:ll}. Let $L(x):= \ell(\eta_\theta, \Phi(x))$, then we can now say that an image $x$ has good quality if 
\begin{align}
    L(x)\sim L\left(\tilde{x}\right)\text{ for }\tilde{x}\in X.
\end{align} 
The advantage of training NFs on hand-crafted properties $\phi^{(i)}$ is that from these features, we already know that they are performance-limiting for the object detection network. Besides, our approach enhances the models' explainability. Additionally, this approach is highly scalable, as properties can be easily added or replaced, since the normalizing flows are relatively small and lightweight neural networks that can be retrained computationally efficiently. 

\subsection{Conversion}
In the next step, we need to align offline testing results with the real-time performance evaluation of the object detectors. Therefore, note that all state-of-the-art object detectors output their predictions together with a confidence score. This additional information indicates for every detected bounding box how certain the network is about its predictions. In inference, a confidence threshold is often set to filter out predictions with low confidence. However, in most AI systems, this threshold is fixed. We propose to couple it with the given image quality. That means, for a low-quality image, we decrease the confidence threshold, emphasizing recall and avoiding the risk of missing any object. For safety-critical systems, such as automated driving, this enhances trust in the system by increasing the likelihood that a pedestrian is detected, even under adverse conditions. In the case of high-quality image data, the emphasis remains on providing the most accurate predictions possible. Motivated by the theory on geometric phase field representations \cite{ModicaMortola1977}, we propose computing the confidence threshold for an image $x$ based on its likelihood of having good quality via the one-dimensional optimal profile for phase transitions. To do this, we divide the space of possible images into two categories: good-quality and poor-quality. The phase transition then describes the process of changing image quality. Therefore, we determine the confidence as 
\begin{align}\label{eq:conf}
    \text{conf}(x) = \frac{1}{2}\tanh\left(\alpha(L(x)-\tau)\right) + \frac{1}{2}.
\end{align}
Here, $\tau\in\mathbb{R}$ is a value that depends on the likelihood distribution of the training dataset. Using this conversion, for an image $x$ with $L(x)=\tau$, the confidence score will be set to $\frac{1}{2}$. We later, in \cref{sec:experiments}, set this to be roughly the median value of the log-likelihood of the train dataset. The value $\alpha > 0$ is a scaling factor, indicating the phase transition width.
% alpha = .3
% return np.clip(np.arctan( alpha * (likelihood - 32.)) / np.pi + .5, a_min=0.0, a_max=.7)
\section{Experiments}
\label{sec:experiments}

 To validate our proposed methodology, we conduct several experiments. First, we train a Real-NVP normalizing flow on the BDD100k training dataset \cite{Yu2018}, an image dataset for automated driving comprising 100,000 images from the US. In a first step, we compute the image qualities of the images based on the features listed in \cref{tab:iqa} and normalize each feature individually. Note, that all subsequential experiments are done with the same flow network. 

\subsection{OOD Detection}
\begin{figure}[H]
  \centering
\includegraphics[width=\columnwidth]{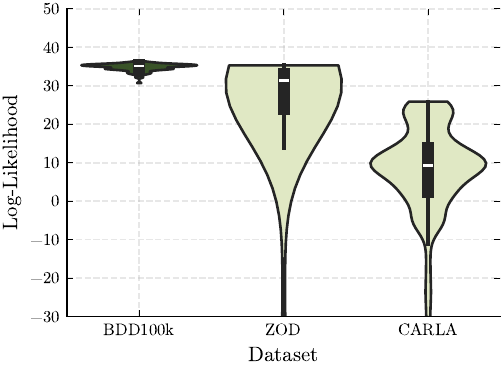} % smaller font
   \caption{Violine plot showing the distribution of image quality $L$ for three different datasets trained on the BDD100k dataset. We randomly sampled 100 images from each dataset. }
   \label{fig:ood}
\end{figure}

First, we demonstrate how to utilize the NF for out-of-distribution detection. To achieve this, we evaluate our network on the BDD100k test dataset and on two additional public datasets: the Zenseact OpenDataset (ZOD) \cite{Alibeigi2023}, containing real-life images for automated driving from Europe, and a dataset comprising simulated images from CARLA \cite{repo_name}. CARLA is a widely used open-source simulator for automated driving. We then apply our likelihood estimation from \cref{al:min} and look at the distribution in \cref{fig:ood}. Here we see evident differences. The BDD100k test data accumulate in an area of high likelihood. This behavior indicates that the network correctly identifies these images as having similar quality properties to those in the BDD100k training dataset. The ZOD dataset shows a broader range. Many images, particularly those above the median, exhibit similar quality characteristics to those in the BDD100k dataset. However, some images differ significantly in quality. Finally, we examine the CARLA data set, which contains simulated data. Here, we observe that it is clearly distinct from the BDD100k data and exhibits minimal overlap. This deviation is because simulated data has fundamentally different image properties than real data, which the NF has recognized.

\subsection{Detection of Faults}
\begin{figure*}
  \centering
   %\includegraphics[width=0.8\linewidth]{egfigure.eps}
\includegraphics[width=.95\linewidth]{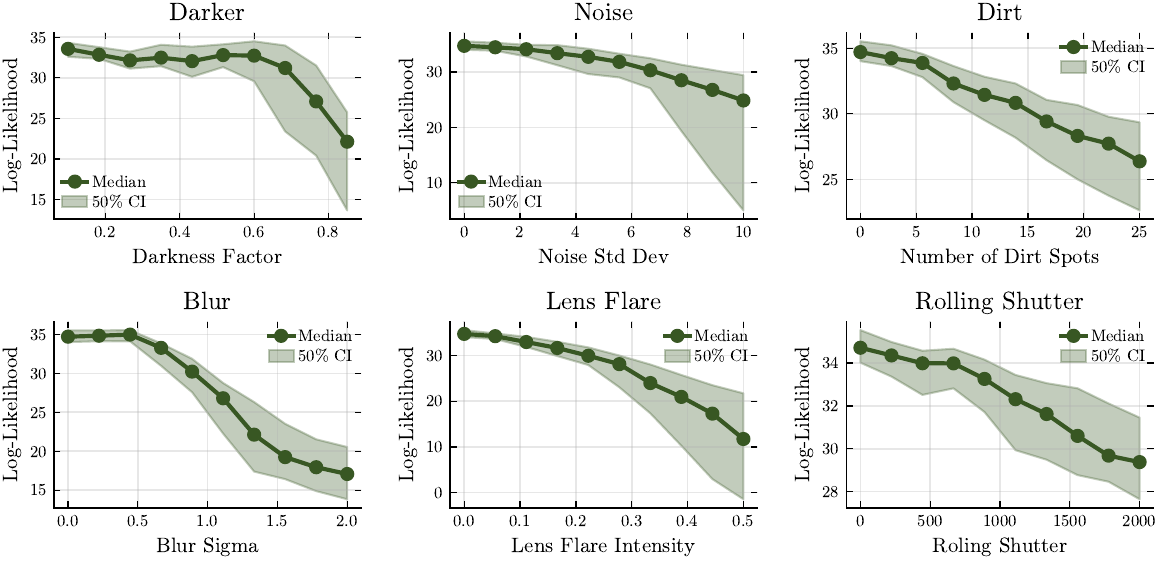}
   \caption{Effect of different image perturbation methods on the logarithmic likelihood of an image having good quality. The $x$-axis shows the intensity of the different perturbations. The green line represents the median value, and the green area indicates the 50\% confidence interval around the median.}
   \label{fig:pertubations}
\end{figure*}
In the next step, we investigate whether the normalizing flow can detect image augmentations. To do this, we select five random images and augment them with 10 different intensity levels. We consider a total of 6 augmentation types:
\begin{itemize}
\item \textbf{Darkness.} We multiply the pixel values by a factor and clip them at the maximum possible pixel value, i.e., 255.
\item \textbf{Noise.} We add Gaussian Noise distributed around zero and increase the standard deviation of the distribution.
\item \textbf{Dirt.} We simulate the effect of dirt on the lens by adding a varying amount of randomly sampled semi-transparent circles of random size and shape to the image.
\item \textbf{Blur.} We blur the image by applying a Gaussian filter and varying the dispersion of the Gaussian curve.
\item \textbf{Lens Flare.} We simulate a lens flare effect as it occurs when strong light falls directly on the camera. We do this by spawning random circles in the image, similar to the dirt effect, which are then superimposed on the image with decreasing intensity. An example can be found at the bottom right in \cref{fig:intro}.
\item \textbf{Rolling shutter.} Here, we simulate a typical misbehavior of a camera sensor with a defective rolling shutter. To do this, we shift each row of pixels horizontally to distort the image.
\end{itemize}

We repeat the partial augmentation 10 times and consider the median values along with a 50\% confidence interval. The results can be seen in \cref{fig:pertubations}. We observe that the normalizing flow is capable of detecting any form of augmentation. The rate at which the flow detects augmentation varies greatly depending on the type of augmentation used. Generally, the spread of quality also tends to increase with the intensity of augmentation. We also repeated the experiment with latent-features from a ResNet-18 instead of our handcraftet ones, but were not able to reliably detect any of the six augmentation types. From that, we conclude that these latent embeddings are indeed not suitable for quality estimates.

\subsection{Combination with Object Detection}
\begin{figure}
  \centering
\includegraphics[width=\columnwidth]{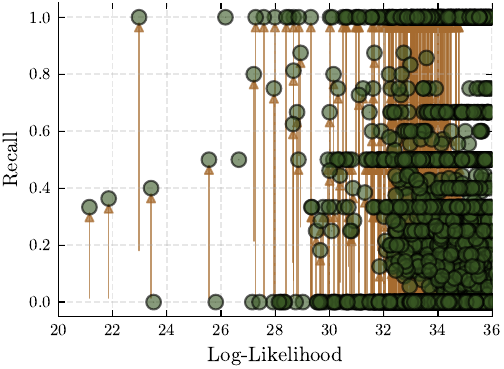} % smaller font
   \caption{Correlation between the logarithmic likelihood of an image having good quality and the recall of the YOLO detector with our fail-degraded system. Orange lines indicate an upward or downward trend, from where the points lie with the constant threshold strategy.}
   \label{fig:cr}
\end{figure}

\begin{figure}
  \centering
\includegraphics[width=\columnwidth]{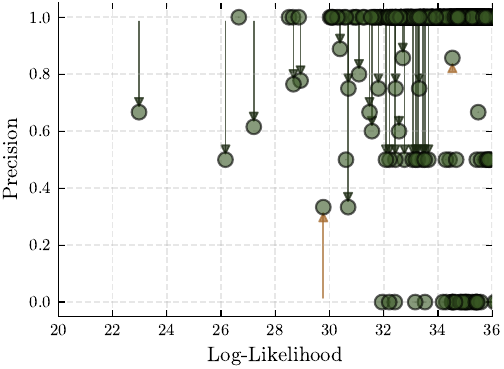} % smaller font
   \caption{Correlation between the logarithmic likelihood of an image having good quality and the precision of the YOLO detector with our fail-degraded system. Orange lines indicate a downward movement, from where the points have been using the constant threshold strategy and violet arrow upward movements.}
   \label{fig:cp}
\end{figure}

In the next step, we explore the integration of the monitor with an object detection algorithm. For this purpose, we train an anchor based YOLOv11 model \cite{yolo11_ultralytics} and a transformer-based DETR \cite{carion2020end} on the BDD100k dataset, but we restrict ourselves to the pedestrian class only. To showcase the connection to the training data, we turn off image augmentations during training; otherwise, we use the default training parameter. For our experiments, we now investigate performance differences across various images. For this, we set an intersection-over-union (IoU) score of 0.5 and calculate the number of true positives, false negatives, and false positives for each image. From that, we also compute precision and recall values for each image individually. For now, we only focus on the YOLO results as the DETR results are quiet similar. First, we fix a confidence threshold of $0.7$ as a reference and consider the initial precision of the individual images in Figure \ref{fig:cp}. 
We notice that precision is high, where the likelihood is low, but we cannot make any statements about images with high likelihood values. The high-precision values mean that if the object detector detects an object in an image of low quality, it is likely to make an accurate prediction. We now also consider the recall in \cref{fig:cr}. 
Here, we observe a downward trend in recall related to image quality. This behaviour indicates that if an image is likely to have low quality, the network tends to miss pedestrians. We now compare this by implementing our fail-degraded system, adjusting the confidence threshold using \cref{eq:conf}, where we set $\tau = 33$ and $\alpha = 0.15$. Additionally, we clip the confidence values above at $0.7$. Examining the overall performance differences of the object detector, we observe that, as expected, our approach yields a decrease in precision and an increase in recall, as shown in \cref{tab:kpi_overview}. \begin{table}
  \caption{Object Detector KPIs}
  \label{tab:kpi_overview}
  \centering
  \begin{tabular}{@{}p{0.3\columnwidth}p{0.17\columnwidth}p{0.05\columnwidth}p{0.17\columnwidth}p{0.05\columnwidth}@{}}
  \hline
   & \multicolumn{2}{c}{YOLO} & \multicolumn{2}{c}{DETR} \\
  \cline{2-5}
   Metric & Fixed Conf. & Ours & Fixed Conf. & Ours \\
  \hline\hline
   Avg. Recall per frame & \centering 0.428 & \centering\textbf{0.474} & \centering 0.605 & \textbf{0.627} \\
   Avg. Precision per frame &\centering \textbf{0.996} & \centering 0.984 &\centering \textbf{0.366} & 0.349 \\
  \hline
  \end{tabular}
\end{table}

Rather than overall statistics, we now investigate further how it affects the individual images. 
We see in \cref{fig:cr} that, as expected, the recall improves significantly for the majority of the low-quality images. This behavior indicates that the number of missed pedestrians is reduced in cases of low-quality images, while maintaining the same recall level for high-quality images. We additionally also consider precision in \cref{fig:cp}. Here, we see that precision on a lot of these images also decreases. By being more cautious with low-quality images, we also detect more pedestrians where there were none. In total, our method improves recall for 377 images, while precision decreases for 25 images and improves for two images.

\subsection{Real-Time Capability}
To demonstrate that our method is truly real-time capable, we provide an implementation based on ROS 2. The Toolchain can be connected to any camera-based image stream. Considering \cref{fig:intro}, all the depicted thick arrows represent a ROS topic, which can then be visualized within standard tools. Note that the forward passes of the object detection algorithm and the normalizing flow are independent of each other, allowing us to run them in parallel. In practice, we have one node assessing the image quality and publishing a confidence value. The object detector computes the forward pass, sets the subscriptions to the confidence value, and outputs bounding box information correctly. 


\section{Conclusion}

To summarize, we began by arguing that not all types of errors are of equal importance for the trustworthiness of a system; for instance, missing a pedestrian is more critical than detecting a ghost. We then argued that for low-quality images, this becomes more crucial as they are more prone to errors. We then defined our fail-degraded system by emphasizing the recall for faulty images. To estimate the quality of images, we trained a normalizing flow network on image statistics and demonstrated how it can detect outliers. This definition of quality indeed correlates with the accuracy of the object detector. We also demonstrated that our fail-degraded system can significantly enhance the recall for low-quality images, while maintaining the same accuracy for high-quality images. Note that this resembles a general trade-off between trustworthiness and performance of AI systems. For high-quality images, the system focuses on giving the most accurate and precise predictions. In cases of low-quality images, we prioritize not missing pedestrians, ensuring we can still trust the perception system to avoid critical mistakes. This behavior also resembles the paradox of automated vehicle safety, which states that the safest automated vehicle is the one that does not drive at all. In our case, for a completely corrupted image, we would focus solely on recall, meaning we would always detect a pedestrian in front of the vehicle. Continuous detection of pedestrians would prevent the vehicle from starting to drive, and thus, the vehicle would never drive. Hence, only knowing that the image contains enough information to make good predictions will lead the AI system to focus more on actually increasing accuracy.

In this work, we focused on global image quality aspects; however, for detecting pedestrians, local aspects may be even more important. In future work, we aim to investigate how to adapt our approach to work locally on bounding boxes and how to incorporate this information into the training process of the object detector. We could do this by emphasizing recall on low-quality objects through asymmetric loss functions.

\section*{Acknowledgements}
{The research leading to these results is funded by the German Federal Ministry for Economic Affairs and Energy within the project "Safe AI Engineering - Sicherheitsargumentation befaehigendes AI Engineering ueber den gesamten Lebenszyklus einer KI-Funktion". The authors would like to thank the consortium for the successful cooperation.}
{
    \small
    \bibliographystyle{IEEEtran}
    \bibliography{main}

@inproceedings{he2016deep,
  title={Deep residual learning for image recognition},
  author={He, Kaiming and Zhang, Xiangyu and Ren, Shaoqing and Sun, Jian},
  booktitle={Proceedings of the IEEE conference on computer vision and pattern recognition},
  pages={770--778},
  year={2016}
}

@inproceedings{caron2021emerging,
  title={Emerging properties in self-supervised vision transformers},
  author={Caron, Mathilde and Touvron, Hugo and Misra, Ishan and J{\'e}gou, Herv{\'e} and Mairal, Julien and Bojanowski, Piotr and Joulin, Armand},
  booktitle={Proceedings of the IEEE/CVF international conference on computer vision},
  pages={9650--9660},
  year={2021}
}

@article{ModicaMortola1977,
  author  = {Modica, Luciano and Mortola, Stefano},
  title   = {Un esempio di {$\Gamma^-$}-convergenza},
  journal = {Bollettino dell'Unione Matematica Italiana},
  series  = {5},
  volume  = {14-B},
  year    = {1977},
  pages   = {285--299},
  language = {it}
}

@inproceedings{carion2020end,
  title={End-to-end object detection with transformers},
  author={Carion, Nicolas and Massa, Francisco and Synnaeve, Gabriel and Usunier, Nicolas and Kirillov, Alexander and Zagoruyko, Sergey},
  booktitle={European conference on computer vision},
  pages={213--229},
  year={2020},
  organization={Springer}
}

@Article{Dinh2016,
  author       = {Laurent Dinh and
                  Jascha Sohl{-}Dickstein and
                  Samy Bengio},
  title        = {Density estimation using Real {NVP}},
  journal      = {CoRR},
  volume       = {abs/1605.08803},
  year         = {2016},
  url          = {http://arxiv.org/abs/1605.08803},
  eprinttype    = {arXiv},
  eprint       = {1605.08803},
  bibsource    = {dblp computer science bibliography, https://dblp.org}
}

@Article{Marchal2019,
  author        = {Marchal, Nicolas and Moraldo, Charlotte and Siegwart, Roland and Blum, Hermann and Cadena, Cesar and Gawel, Abel},
  journal       = {IEEE Robotics and Automation Letters, vol. 5, no. 2, pp. 1032-1038, April 2020},
  title         = {Learning Densities in Feature Space for Reliable Segmentation of Indoor Scenes},
  year          = {2019},
  issn          = {2377-3774},
  month         = apr,
  number        = {2},
  pages         = {1032--1038},
  volume        = {5},
  archiveprefix = {arXiv},
  copyright     = {arXiv.org perpetual, non-exclusive license},
  date          = {2019-08-01},
  doi           = {10.1109/lra.2020.2967313},
  eprint        = {1908.00448},
  primaryclass  = {cs.CV},
  publisher     = {Institute of Electrical and Electronics Engineers (IEEE)},
}

@Article{Kirichenko2020,
   title={Why normalizing flows fail to detect out-of-distribution data},
  author={Kirichenko, Polina and Izmailov, Pavel and Wilson, Andrew G},
  journal={Advances in neural information processing systems},
  volume={33},
  pages={20578--20589},
  year={2020}
}

@Article{Cook2024,
    title={Feature density estimation for out-of-distribution detection via normalizing flows},
  author={Cook, Evan D and Lavoie, Marc-Antoine and Waslander, Steven L},
  journal={arXiv preprint arXiv:2402.06537},
  year={2024}
}

@Article{Fursa2021,
    title={Worsening perception: Real-time degradation of autonomous vehicle perception performance for simulation of adverse weather conditions},
  author={Fursa, Ivan and Fandi, Elias and Musat, Valentina and Culley, Jacob and Gil, Enric and Teeti, Izzeddin and Bilous, Louise and Sluis, Isaac Vander and Rast, Alexander and Bradley, Andrew},
  journal={arXiv preprint arXiv:2103.02760},
  year={2021}
}

@Article{Li2022,
    title={Emergent visual sensors for autonomous vehicles},
  author={Li, You and Moreau, Julien and Ibanez-Guzman, Javier},
  journal={IEEE Transactions on Intelligent Transportation Systems},
  volume={24},
  number={5},
  pages={4716--4737},
  year={2023},
  publisher={IEEE}
}

@Article{Zhang2023,
  author    = {Zhang, Yuxiao and Carballo, Alexander and Yang, Hanting and Takeda, Kazuya},
  journal   = {ISPRS Journal of Photogrammetry and Remote Sensing},
  title     = {Perception and sensing for autonomous vehicles under adverse weather conditions: A survey},
  year      = {2023},
  issn      = {0924-2716},
  month     = feb,
  pages     = {146--177},
  volume    = {196},
  doi       = {10.1016/j.isprsjprs.2022.12.021},
  publisher = {Elsevier BV},
}

@Article{Mittal2012,
  author    = {Mittal, A. and Moorthy, A. K. and Bovik, A. C.},
  journal   = {IEEE Transactions on Image Processing},
  title     = {No-Reference Image Quality Assessment in the Spatial Domain},
  year      = {2012},
  issn      = {1941-0042},
  month     = dec,
  number    = {12},
  pages     = {4695--4708},
  volume    = {21},
  doi       = {10.1109/tip.2012.2214050},
  publisher = {Institute of Electrical and Electronics Engineers (IEEE)},
}

@Article{Mittal2013,
  author    = {Mittal, A. and Soundararajan, R. and Bovik, A. C.},
  journal   = {IEEE Signal Processing Letters},
  title     = {Making a “Completely Blind” Image Quality Analyzer},
  year      = {2013},
  issn      = {1558-2361},
  month     = mar,
  number    = {3},
  pages     = {209--212},
  volume    = {20},
  doi       = {10.1109/lsp.2012.2227726},
  publisher = {Institute of Electrical and Electronics Engineers (IEEE)},
}

@inproceedings{Liu2017,
    title={Rankiqa: Learning from rankings for no-reference image quality assessment},
  author={Liu, Xialei and Van De Weijer, Joost and Bagdanov, Andrew D},
  booktitle={Proceedings of the IEEE international conference on computer vision},
  pages={1040--1049},
  year={2017}
}

@Article{Kim2019,
  author    = {Kim, Jongyoo and Nguyen, Anh-Duc and Lee, Sanghoon},
  journal   = {IEEE Transactions on Neural Networks and Learning Systems},
  title     = {Deep CNN-Based Blind Image Quality Predictor},
  year      = {2019},
  issn      = {2162-2388},
  month     = jan,
  number    = {1},
  pages     = {11--24},
  volume    = {30},
  doi       = {10.1109/tnnls.2018.2829819},
  publisher = {Institute of Electrical and Electronics Engineers (IEEE)},
}

@InProceedings{Huang2024,
  author    = {Huang, Yijie and Ni, Haoyang and Zhang, Kaiwei and Jia, Ziheng and Lu, Fangfang and Min, Xiongkuo and Zhai, Guangtao},
  booktitle = {2024 IEEE International Conference on Visual Communications and Image Processing (VCIP)},
  title     = {ACIQA: A Dataset and Method for Assessing the Imaging Quality of Automotive Cameras},
  year      = {2024},
  month     = dec,
  pages     = {1--5},
  publisher = {IEEE},
  doi       = {10.1109/vcip63160.2024.10849918},
}

@InProceedings{RamirezAgudelo2025,
  author    = {Ramírez-Agudelo, Oscar H. and Gorea, Nicoleta and Reif, Aliza and Bonasera, Lorenzo and Karl, Michael},
  booktitle = {Applications of Machine Learning 2025},
  title     = {The role of noisy data in improving CNN robustness for image classification},
  year      = {2025},
  editor    = {Narayanan, Barath and Zelinski, Michael E. and Taha, Tarek M. and Awwal, Abdul A. and Iftekharuddin, Khan M.},
  month     = sep,
  pages     = {27},
  publisher = {SPIE},
  doi       = {10.1117/12.3063563},
}

@InProceedings{Lyssenko2022,
  author    = {Lyssenko, Maria and Gladisch, Christoph and Heinzemann, Christian and Woehrle, Matthias and Triebel, Rudolph},
  booktitle = {2022 IEEE/RSJ International Conference on Intelligent Robots and Systems (IROS)},
  title     = {Towards Safety-Aware Pedestrian Detection in Autonomous Systems},
  year      = {2022},
  month     = oct,
  pages     = {293--300},
  publisher = {IEEE},
  doi       = {10.1109/iros47612.2022.9981309},
}

@InProceedings{Feifel2022,
            year = {2022},
           month = {Juli},
          author = {Feifel, Patrick and Bonarens, Frank and Franke, Benedikt and Raulf, Arne Peter and K{\"o}ster, Frank and Schwenker, Friedhelm},
       booktitle = {2022 Workshop on Artificial Intelligence Safety, AISafety 2022},
           title = {Revisiting the Evaluation of Deep Neural Networks for Pedestrian Detection},
             url = {https://elib.dlr.de/195574/}
}

@Article{Yang2022,
   title={Image data augmentation for deep learning: A survey},
  author={Yang, Suorong and Xiao, Weikang and Zhang, Mengchen and Guo, Suhan and Zhao, Jian and Shen, Furao},
  journal={arXiv preprint arXiv:2204.08610},
  year={2022}
}

@misc{yolo11_ultralytics,
  author = {Glenn Jocher and Jing Qiu},
  title = {Ultralytics YOLO11},
  version = {11.0.0},
  year = {2024},
  howpublished = {https://github.com/ultralytics/ultralytics},
  orcid = {0000-0001-5950-6979, 0000-0003-3783-7069},
  license = {AGPL-3.0}
}

@misc{repo_name,
  author       = {DanielHfnr},
  title        = {Carla-Object-Detection-Dataset},
  year         = {2023},
  howpublished          = {https://github.com/DanielHfnr/Carla-Object-Detection-Dataset},
  note         = {Accessed: 2025-11-02},
}

@InProceedings{Venkatanath2015,
  author    = {Venkatanath N and Praneeth D and Maruthi Chandrasekhar Bh and Channappayya, Sumohana S. and Medasani, Swarup S.},
  booktitle = {2015 Twenty First National Conference on Communications (NCC)},
  date      = {2015-02},
  year         = {2015},
  title     = {Blind image quality evaluation using perception based features},
  doi       = {10.1109/ncc.2015.7084843},
  pages     = {1--6},
  publisher = {IEEE},
}

@Article{Nalisnick2018,
    title={Do deep generative models know what they don't know?},
  author={Nalisnick, Eric and Matsukawa, Akihiro and Teh, Yee Whye and Gorur, Dilan and Lakshminarayanan, Balaji},
  journal={arXiv preprint arXiv:1810.09136},
  year={2018}
}

@article{yurtsever2020survey,
  title={A survey of autonomous driving: Common practices and emerging technologies},
  author={Yurtsever, Ekim and Lambert, Jacob and Carballo, Alexander and Takeda, Kazuya},
  journal={IEEE access},
  volume={8},
  pages={58443--58469},
  year={2020},
  publisher={IEEE}
}

@Article{Lee2024,
   title={Toward robust LiDAR based 3D object detection via density-aware adaptive thresholding},
  author={Lee, Eunho and Jung, Minwoo and Kim, Ayoung},
  journal={arXiv preprint arXiv:2404.13852},
  year={2024}
}

@inproceedings{Alibeigi2023,
  title={Zenseact open dataset: A large-scale and diverse multimodal dataset for autonomous driving},
  author={Alibeigi, Mina and Ljungbergh, William and Tonderski, Adam and Hess, Georg and Lilja, Adam and Lindstr{\"o}m, Carl and Motorniuk, Daria and Fu, Junsheng and Widahl, Jenny and Petersson, Christoffer},
  booktitle={Proceedings of the IEEE/CVF International Conference on Computer Vision},
  pages={20178--20188},
  year={2023}
}

@InProceedings{Hasler2003,
author    = {Hasler, David and Suesstrunk, Sabine E.},
booktitle = {Human Vision and Electronic Imaging VIII},
title     = {Measuring colorfulness in natural images},
editor    = {Rogowitz, Bernice E. and Pappas, Thrasyvoulos N.},
publisher = {SPIE},
date      = {2003-06},
year         = {2003},
doi       = {10.1117/12.477378},
issn      = {0277-786X},
}

@inproceedings{rezende2015variational,
  title={Variational inference with normalizing flows},
  author={Rezende, Danilo and Mohamed, Shakir},
  booktitle={International conference on machine learning},
  pages={1530--1538},
  year={2015},
  organization={PMLR}
}

@article{schmid2021,
  title={Formal Verification of a Fail-Operational Automotive Driving System},
  author={Schmid, Tobias and Schraufstetter, Stefanie and Fritzsch, Jonas and Hellhake, Dominik and Koelln, Greta and Wagner, Stefan},
  journal={arXiv preprint arXiv:2101.07307},
  year={2021}
}

@inproceedings{Yu2018,
  title={Bdd100k: A diverse driving dataset for heterogeneous multitask learning},
  author={Yu, Fisher and Chen, Haofeng and Wang, Xin and Xian, Wenqi and Chen, Yingying and Liu, Fangchen and Madhavan, Vashisht and Darrell, Trevor},
  booktitle={Proceedings of the IEEE/CVF conference on computer vision and pattern recognition},
  pages={2636--2645},
  year={2020}
}

@Article{Stolte2022,
  author       = {Stolte, Torben and Ackermann, Stefan and Graubohm, Robert and Jatzkowski, Inga and Klamann, Bjorn and Winner, Hermann and Maurer, Markus},
  date         = {2022-06},
  journal = {IEEE Transactions on Intelligent Vehicles},
  title        = {Taxonomy to Unify Fault Tolerance Regimes for Automotive Systems: Defining Fail-Operational, Fail-Degraded, and Fail-Safe},
  doi          = {10.1109/tiv.2021.3129933},
  issn         = {2379-8858},
  number       = {2},
  pages        = {251--262},
  year         = {2022},
  volume       = {7},
  publisher    = {Institute of Electrical and Electronics Engineers (IEEE)},
}

@misc{weiss2023reliabilityanalysisgracefullydegrading,
      title={Reliability Analysis of Gracefully Degrading Automotive Systems}, 
      author={Philipp Weiss and Ali Younessi and Sebastian Steinhorst},
      year={2023},
      eprint={2305.07401},
      archivePrefix={arXiv},
      primaryClass={cs.DC},
      url={https://arxiv.org/abs/2305.07401}, 
}

@misc{valdenegrotoro2022deeperlookaleatoricepistemic,
      title={A Deeper Look into Aleatoric and Epistemic Uncertainty Disentanglement}, 
      author={Matias Valdenegro-Toro and Daniel Saromo},
      year={2022},
      eprint={2204.09308},
      archivePrefix={arXiv},
      primaryClass={cs.LG},
      url={https://arxiv.org/abs/2204.09308}, 
}

@techreport{tuevwhitepaper,
  author      = {Benjamin Koller, Fabian Frey, Kilian Zwirglmaier},
  title       = {Safety in ADAS/AD –SOTIF, a risk-based approach},
  institution = {TÜV SÜD AG, Qualcomm Technologies, Inc.},
  year        = {2023},
  type        = {Whitepaper},
  number      = {TR-2023-04},
  url         = {https://www.tuvsud.com/-/jssmedia/global/pdf-files/whitepaper-report-e-books/tuvsud-sotif.pdf}
}

@Article{Hossam2025,
  author       = {Hossam, Abdallah and Villagra, Jorge and Navas, Francisco and Milanés, Vicente},
  year         = {2025},
  journal = {IEEE Transactions on Intelligent Transportation Systems},
  title        = {Scalable Fail-Degraded Systems for Autonomous Vehicles: A Survey},
  doi          = {10.1109/tits.2025.3612262},
  issn         = {1558-0016},
  pages        = {1--19},
  publisher    = {Institute of Electrical and Electronics Engineers (IEEE)},
}

@techreport{kia2022,
  author      = {Gemeinsamer Abschlussberichts des Verbundprojektes KI Absicherung},
  institution = {VDA Leitinitiative autonomes und vernetztes Fahren},
  title       = {Abschlussbericht KI Absicherung},
  year        = {2022},
  url         = {https://www.ki-absicherung-projekt.de/fileadmin/KI_Absicherung/Downloads/KI-A_Abschlussberich_PU_final.pdf}
}
}

%WARNING: do not forget to delete the supplementary pages from your submission 

\end{document}